\def\year{2019}\relax
\documentclass[letterpaper]{article} 
\usepackage{aaai19}  
\usepackage{times}  
\usepackage{helvet}  
\usepackage{courier}  
\usepackage{url}  
\usepackage{graphicx}  
\usepackage{amsmath}
\usepackage{amsfonts}
\usepackage{xcolor}
\usepackage{tabularx}

\title{Pathological Evidence Exploration in Deep Retinal Image Diagnosis}
\author{
Yuhao Niu,\textsuperscript{1,2,}\thanks{These two authors contributed equally to the paper.}
Lin Gu,\textsuperscript{4,}$^*$
Feng Lu,\textsuperscript{1,2,3,}\thanks{Corresponding Author: Feng Lu (lufeng@buaa.edu.cn)}
Feifan Lv,\textsuperscript{1,3}
Zongji Wang,\textsuperscript{1}
Imari Sato,\textsuperscript{4}\\{\bf \Large 
Zijian Zhang,\textsuperscript{5}
Yangyan Xiao,\textsuperscript{6}
Xunzhang Dai,\textsuperscript{5}
Tingting Cheng\textsuperscript{5}
} \\
\textsuperscript{1}State Key Laboratory of VR Technology and Systems, School of CSE, Beihang University, Beijing, China \\
\textsuperscript{2}Beijing Advanced Innovation Center for Big Data-Based Precision Medicine, Beihang University, Beijing, China\\
\textsuperscript{3}Peng Cheng Laboratory, Shenzhen, China \quad
\textsuperscript{4}National Institute of Informatics, Japan \\
\textsuperscript{5}Xiangya Hospital Central South University, China \quad
\textsuperscript{6}The Second Xiangya Hospital of Central South University, China
}

\begin{document}
	%
	
	\maketitle
	\begin{abstract}
		Though deep learning has shown successful performance in classifying the label and severity stage of certain disease, most of them give few evidence on how to make prediction. Here, we propose to exploit the interpretability of deep learning application in medical diagnosis. Inspired by Koch's Postulates, a well-known strategy in medical research to identify the property of pathogen, we define a pathological descriptor that can be extracted from the activated neurons of a diabetic retinopathy detector. To visualize the symptom and feature encoded in this descriptor, we propose a GAN based method to synthesize pathological retinal image given the descriptor and a binary vessel segmentation. Besides, with this descriptor, we can arbitrarily manipulate the position and quantity of lesions. As verified by a panel of 5 licensed  ophthalmologists, our synthesized images carry the symptoms that are directly related to diabetic retinopathy diagnosis. The panel survey also shows that our generated images is both qualitatively and quantitatively superior to existing methods. 
	\end{abstract}
	
	\section{Introduction}
	
	Deep learning has become a popular methodology in analyzing medical images such as diabetic retinopathy detection~\cite{Gulshan2016jama}, classifying skin cancer~\cite{esteva2017nature}. Though these algorithms have proven quite accurate in classifying specific disease label and severity stage, most of them lack the ability to explain its decision, a common problem that haunts deep learning community. Lacking interpretability is especially imperative for medical image application, as physicians or doctors relies on medical evidence to determine whether to trust the machine prediction or not.
	
	In this paper, we propose a novel technique inspired by  Koch's Postulates to give some insights into how convolutional neural network (CNN) based pathology detector makes decision. In particular, we take the diabetic retinopathy detection network~\cite{oO2016detector} for example. Noted that not limited for ~\cite{oO2016detector}, this strategy could also be extended to interpret more general deep learning model.
    
\begin{figure*}[h!]
		\begin{center}
			\includegraphics[scale=0.52]{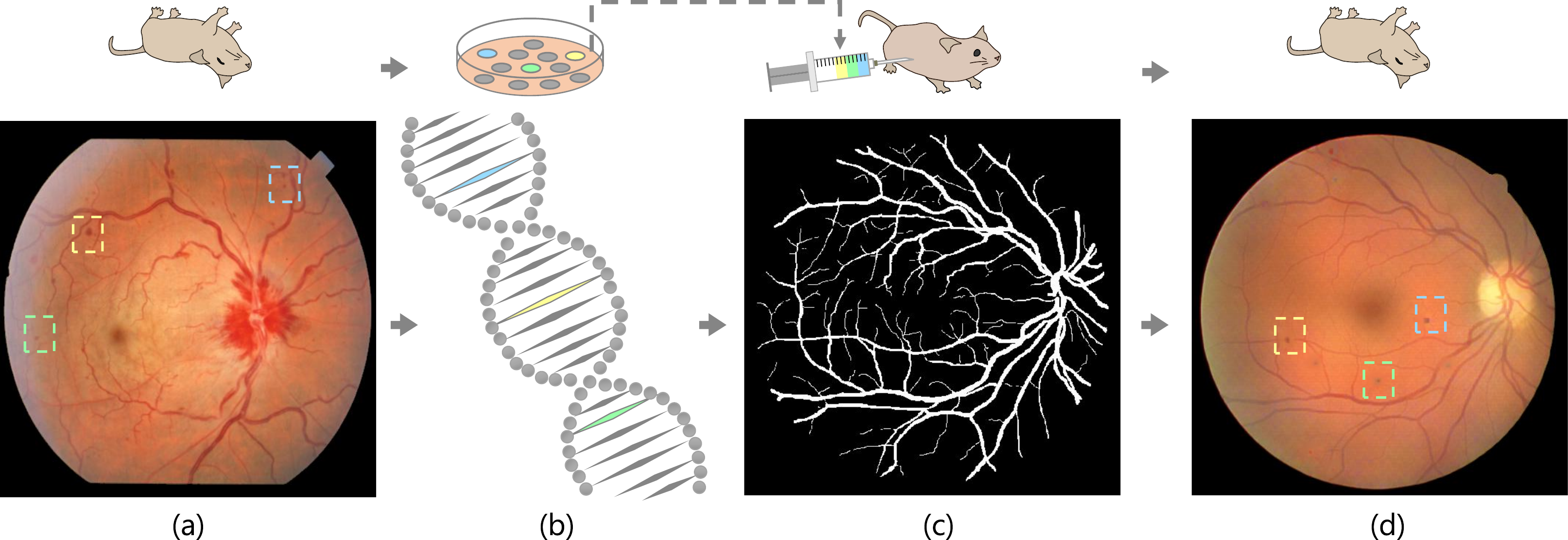}
		\end{center}
		\caption{Koch's Postulates are criteria in Evidence Based Medicine (EBM) to determine the pathogen for a certain disease. They state that the pathogen must be found in diseased subjects but not in healthy ones; the pathogen must be isolated and grown in pure culture; the cultured pathogen should cause disease after injected into healthy subject; the pathogen isolated again is the same as the injected one. The methodology of this paper is an analogy to Koch's Postulates. (a) Reference retinal image with disease. (b) Extract pathological descriptor from an image like separating pathogen. (c) Apply the descriptor on a binary vessel segmentation like injecting purified pathogen into the subject. (d) The synthesized image or subject with same symptom.}
		\label{fig:koch}
	\end{figure*}

    We at first apply~\cite{oO2016detector} on the  reference image (Fig~\ref{fig:koch}.(a)) and extract the pathological descriptor (Fig~\ref{fig:koch}.(b)) that encodes the neuron activation directly related to prediction. Picking thousand out of  millions of parameters in neuron network is like separating the potential pathogen. Koch's Postulates claims that the property of pathogen, though invisible for naked eye, could be determined by observing the arose symptom after injecting it into subject.  Similarly, we \emph{inject} the pathologic descriptor into the binary vessel segmentation (Fig~\ref{fig:koch}.(c)) to synthesize the retinal image. We achieve this with a GAN based network as illustrated in Fig~\ref{fig:pipeline}. Given pathologic descriptor and binary vessel segmentation, our generated image (Fig~\ref{fig:koch}.(d))  exhibits the expected symptom such as microaneurysms and hard exudates that appear in the target image. Since our descriptor is lesion-based and spatial independent, we could arbitrarily manipulate the position and number of lesions. Evaluated With a panel of 5  licensed  ophthalmologists, our generated retinal images are  qualitatively and quantitatively superior to existing methods.



	Specifically, we  encode a series of pathological descriptor as illustrated in Fig~\ref{fig:descriptor}. According to our analysis, the diabetic retinopathy detection network~\cite{oO2016detector} predicts the diabetic level through a few dimensions (6 among 1024, lighted with colors) of bottleneck feature in Fig~\ref{fig:descriptor}. We then identify the neurons that directly contribute to these 6 dimension bottleneck features in the activation net and record their position and activation value as the pathologic descriptor. Since neuron activation is spatially correlated with individual lesion, our descriptor is defined as lesion-based that allows us to manipulate its position and quantity.
	

	Our main contributions are mainly three-folds:
\begin{enumerate}
\item We define a pathological descriptor that encodes the key parameters of CNN which is directly related to disease prediction. This descriptor is associated with individual lesion. 
\item Inspired by Koch's Postulates, we propose a novel interpretability strategy to visualize the pathological descriptor by synthesizing fully controllable pathological images. The synthesized images are  verified  by a group of licensed  ophthalmologists.
\item With our pathological descriptors, we could generate medical plausible pathology retinal image where the position and quantity of lesion could arbitrarily manipulated.
\end{enumerate}

	\section{Related Works}
    \subsection{Diabetic Retinopathy  Detection}
    \label{sec:DR detector}
With the fast development of deep learning, this technique has achieved success on several medical image analysis applications, such as computer-aided diagnosis of skin cancer, lung node, breast cancer, \textit{etc.} In the case of diabetic retinopathy (DR), automatic detection is particularly needed to reduce the workload of ophthalmologists, and slow down the progress of DR by performing early diagnose on diabetic patients~\cite{Gulshan2016jama}.

     In 2015, a Kaggle competition~\cite{kaggle2016diabetic} was organized to automatically classify retinal images into five stages according to \textit{International Clinical Diabetic Retinopathy Disease Severity Scale}~\cite{aao2002drscale}. Not surprisingly, all of the top-ranking methods were based on deep learning. Then, another deep learning method~\cite{Gulshan2016jama}, trained on  128 175 images, achieved a high sensitivity and specificity for detecting diabetic retinopathy. Noting that image-level grading lacks intuitive explanation, the recent methods~\cite{Yang17MICCAI,wang2017zoom} shifted the focus to locate the lesion position. However, these methods often relied on a large training set of lesion annotations from professional experts.
     
     In this paper, we propose a novel strategy to encode the descriptor from the DR detector's activated neurons directly related to the pathology. For the sake of generality, we select the o\_O\cite{oO2016detector}, a CNN based method within the top-3 entries on Kaggle's challenge. Even now, the performance of o\_O is still equivalent to the latest method~\cite{wang2017zoom}. This method is trained and tested on the image level DR severity stage.



	

   
	\subsection{Generative Adversarial Networks}
Generative Adversarial Networks (GANs)~\cite{goodfellow2014generative} were first proposed in 2014, adopting the idea of zero-sum game. Subsequently, CGANs~\cite{mirza2014conditional} attempted to use additional information to make the GAN controllable. DCGAN~\cite{radford2015unsupervised} combined CNN with traditional GAN to achieve a shocking effect. Pix2pix~\cite{isola2017image} used the U-Net~\cite{ronneberger2015u} combined with adversarial training and achieved amazing results. CycleGAN~\cite{zhu2017unpaired} used two sets of GANs and added cycle loss to achieve style transfer on unpaired data.

    		\begin{figure*}[h!]
		\begin{center}
			\includegraphics[width=\textwidth]{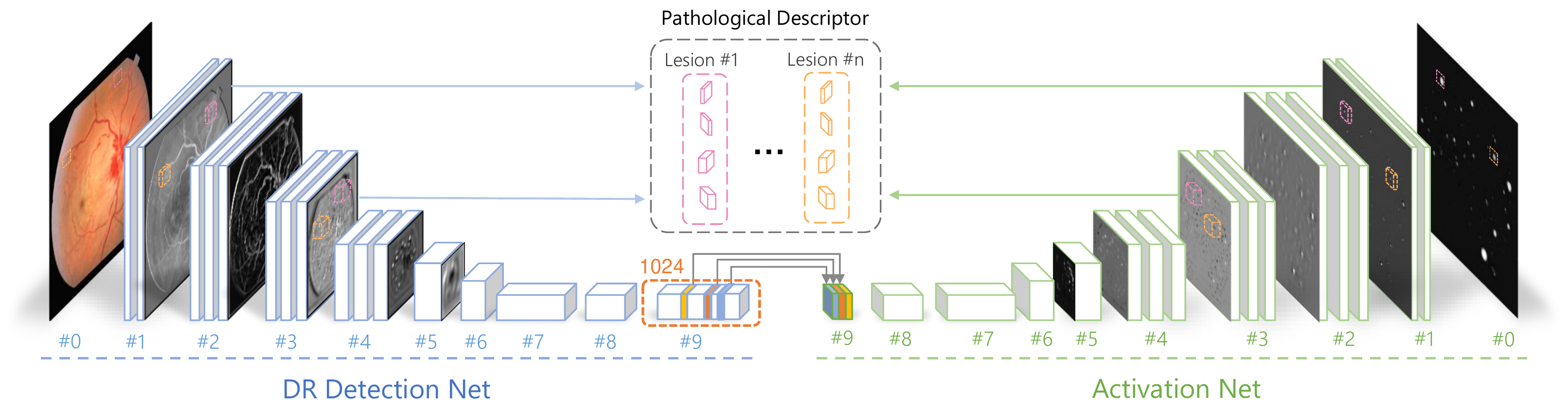}
		\end{center}
		\caption{The process for extracting pathological descriptors. First, a pathological reference image is fed into the DR detection net. Next, the extracted features are mapped to the input pixel space through the activation net to get activation projections, which indicate the locations and appearance of most lesions. Finally, the features and related activation projections are cropped into small patches around the found lesions, which are recognized as pathological descriptors. }
		\label{fig:descriptor}
	\end{figure*}

	\subsection{Style Transfer}
    Recently, neural style transfer using deep CNNs becomes popular. A typical method was proposed in \cite{DBLP:conf/cvpr/GatysEB16}, in which they directly optimized the input pixels to restrict both content and style features extracted by CNNs. 
    Later in \cite{DBLP:conf/cvpr/LuanPSB17} and \cite{DBLP:conf/cvpr/GatysEBHS17}, semantic masks were introduced to improve image style transfer.
	To speed up the transfer procedure, \cite{DBLP:conf/eccv/JohnsonAF16} added a network to synthesize image. Once trained by a given style reference image, it can finish style transfer by one feed-forward propagation. 

	\subsection{Synthesizing Biomedical Images}

	The traditional biomedical imaging synthesizing uses the medical and biological prior knowledge accumulated by humans, combined with complex simulation methods to produce realistic results. 	Probably most well-known efforts are the work~\cite{fiorini2014automatic}, the work~\cite{bonaldi2016automatic}, GENESIS~\cite{bower2015genesis}, NEURON~\cite{carnevale2006neuron}, L-Neuron~\cite{ascoli2000neuron} \textit{etc.}
	With the development of deep learning, some methods like~\cite{zhao2018synthesizing} began to synthesize realistic retinal and neuronal images in a data-driven way. Tub-sGAN, a variant of~\cite{zhao2018synthesizing}, synthesized image given a binary tubular annotation and a reference image. Although their generated images could show pleasant visual appearance, the diabetic retinopathy symptoms and retina physiological details are either lost or  incorrect as verified by the ophthalmologists. In this paper, we propose a pathologically controllable method that can generate realistic retinal image with medical plausible symptoms.

\section{Pathological Descriptor}

In this section, we would describe how to extract lesion based pathological descriptor from the activated neurons of Diabetic Retinopathy (DR) detector~\cite{oO2016detector}.

  

	
	
	\subsection{Diabetic Retinopathy Detection}
  Here, we  briefly introduces  DR detector~\cite{oO2016detector} used in this paper. It takes retinal fundus image of shape $448 \times 448 \times 3$ as input  and outputs the 5 grades (0-4) diabetic retinopathy severity.
   As shown in the left part of Fig~\ref{fig:descriptor}, the DR detection network is stacked with several blocks, each of which consists of 2-3 convolutional layers and a pooling layer. As the number of layers increases, the network merges into a  $ 1\times 1 \times 1024$ bottleneck feature. To add nonlinearity to the net and to avoid neuronal death, a leaky ReLU~\cite{maas2013rectifier} with negative slope 0.01 is applied following each convolutional and dense layer.
   
   The bottleneck feature is fed into a dense layer (not shown in the figure) to predict the severity labels provided in the \textit{DR Detection Challenge} \cite{kaggle2016diabetic}. The network is trained with Nesterov momentum over 250 epochs. Data augmentation methods, such as dynamic data re-sampling, random stretching, rotation, flipping, and color augmentation, are all applied. We refered to ~\cite{oO2016detector} for details.

    \subsection{Key Bottleneck Features}
    
   We further identify a few \emph{key features} (colored one in the middle of Fig~\ref{fig:descriptor}) from $1024$ dimension bottleneck features.  After the training stage of network, we are able to generate the bottleneck features for individual sample in the training set. Then we train a random forest~\cite{Dollar15PAMI} classifier to predict the severity label on these bottleneck features. Following~\cite{Gu17MICCAI}, we could identify the important features by counting the frequency of each feature that contributes to prediction. In the current setting, we find that, with random forest, only $6$  of  $1024$ bottleneck  features could deliver equivalent performance as o\_O~\cite{oO2016detector}.

	
	
	\subsection{Activation Network}
    
    Among millions of neurons in the network, only thousands of neurons actually contribute to the bottleneck feature's activation and the final prediction. To explore the activity of these neurons, we introduce an activation network~\cite{zeiler2014visualizing}. We perform a back-propagation-liked procedure from the $6$ dimension key bottleneck features to get \emph{activation projections} for detector feature layers.

    As shown in the right part of Fig~\ref{fig:descriptor}, our activation net is a reverse version of the DR detector. For individual layer in detector, there is a corresponding reverse layer with the same configuration of strides and kernel size. For a convolutional layer, the corresponding layer performs transposed convolution, which shares same weights, except that the kernel is flipped vertically and horizontally. For each max pooling layer, there is an unpooling layer that conducts a partially inverse operation, where the max elements are located through a skip connection and non-maximum elements are filled with zeros. For a ReLU function, there is also a ReLU in the activation net, which drops out the negative projection. We treat the fully connection as $1\times 1$ convolution. In the implementation we use auto-differentiation provided in Tensorflow~\cite{abadi2016tensorflow} to reverse each layer. 
   
	\begin{figure}[t]
		\begin{center}
			\includegraphics[width=\columnwidth]{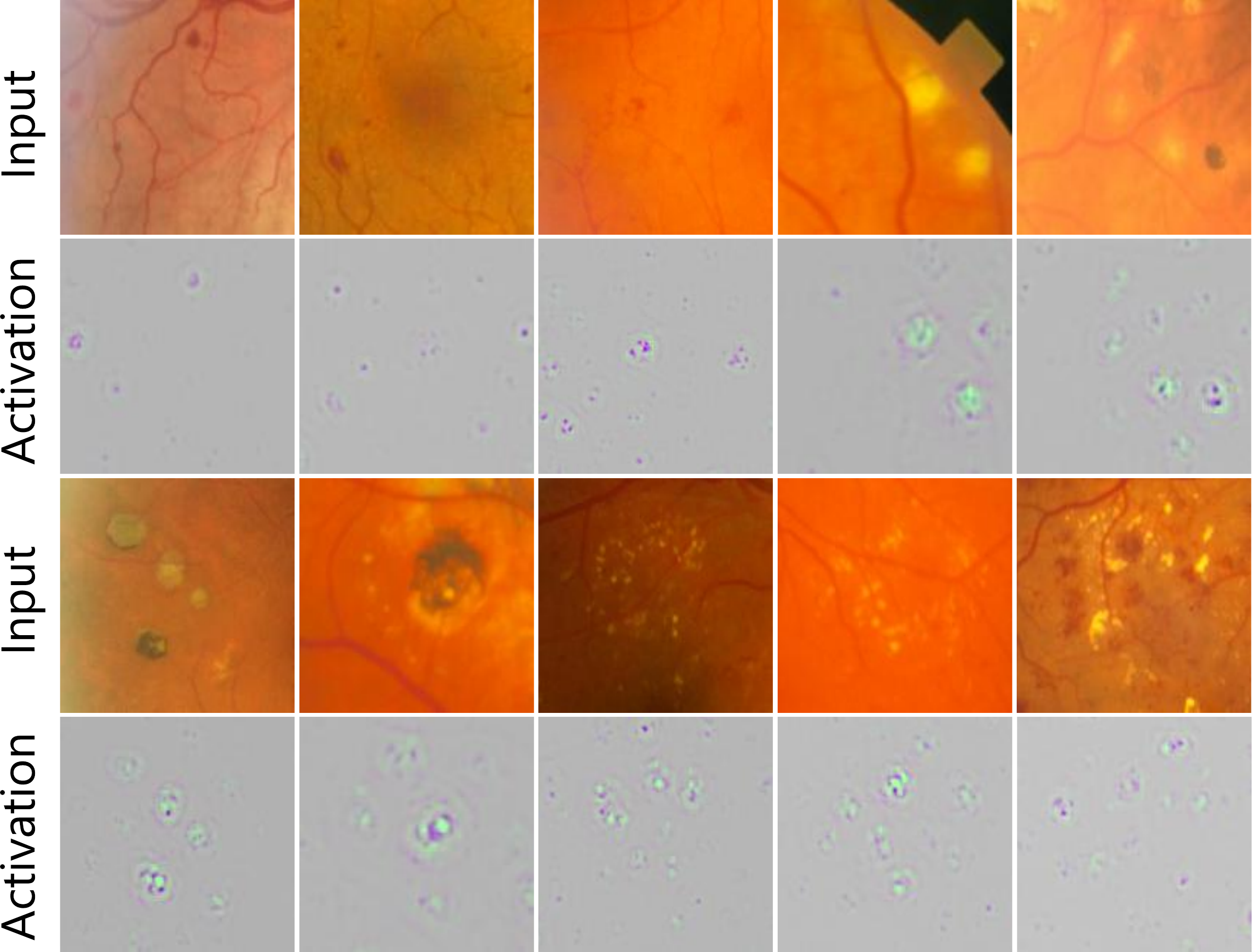}
		\end{center}
		\caption{We input fundus with different lesions into the pipeline in Fig~\ref{fig:descriptor} and extract their related activation projections in layer \#0. Some results are cropped and shown here. }
		\label{fig:detected_lesion}
	\end{figure}
	
	\begin{figure}[t]
		\begin{center}
			\includegraphics[width=\columnwidth]{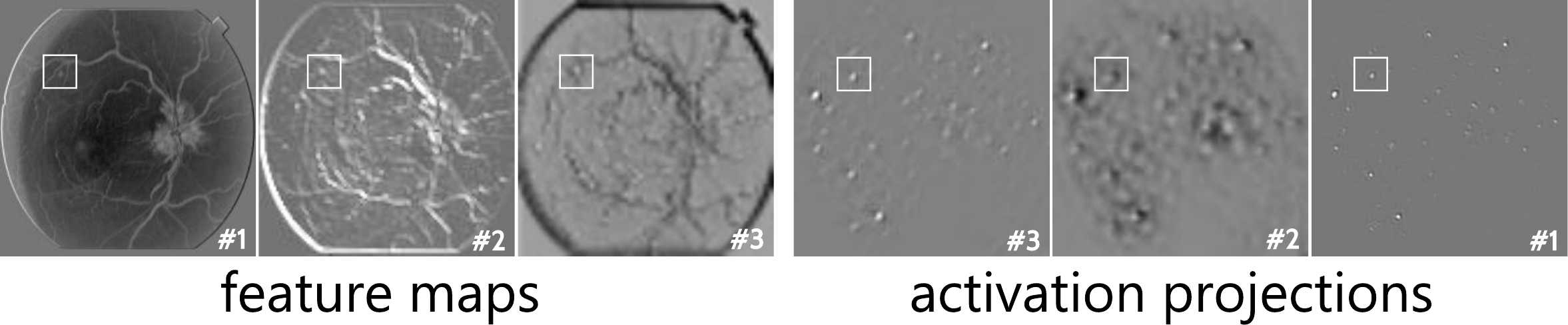}
		\end{center}
		\caption{Feature maps and related activation projections. In each layer, only one channel is shown. }
		\label{fig:feat_proj}
	\end{figure}
	
    Here, we demonstrate some neuron activation examples in Fig~\ref{fig:detected_lesion}\&\ref{fig:feat_proj}. It shows that, though our DR detector is trained with the image-level labels,
    its neuronal activity is sensitive to and threfore can locate a variety of DR lesions such as microaneurysms, soft exudates and hard exudates.
    
	
	\subsection{Retinal Pathological Descriptor}
    
    Now, we define a \emph{pathological descriptor} to encode lesion feature and activation, which should serve as the evidence when doctors make diagnose. As the neuron activation is spatially correlated with the retinal lesions, we could associate the descriptor with individual lesion.  Fig~\ref{fig:feat_proj} shows how the lesion feature changes across different layers. A descriptor contains patches cropped from these maps.
   
	
	When a retinal fundus $ x_s $ is fed into the pipeline in Fig~\ref{fig:descriptor}, the features and activation projections in layer $ l $ are denoted as $ F_l $ and $ A_l $, respectively. In order to indicate position and boundary of an individual lesion, the last layer activation $ A_0 $ is thresholded into a binary mask $ M_0 $. To describe a lesion area, we define a rectangle region $r$ that covers a connected conponent (a lesion) in $ M_0 $. Thus different lesions could be denoted as different $r$. We then use the multi-layer information to construct our retinal pathological descriptor for each lesion. In particular, we first down-sample $M_0$ into each layer $l$ to generate the binary lesion mask $M_l$. With $r$, we cropped feature patch $F_{lr}$ from $F_l$, activation patch $A_{lr}$ from $A_l$ and mask patch $M_{lr}$ from $M_l$. The pathological descriptor for lesion $r$ consists of the information from multiple layers, written as $d = \{d_{l}|l\in \Lambda\}$, where $ d_{l} = \left<M_{lr}, A_{lr}, F_{lr}\right> $.


	
	\begin{figure*}[h!]
		\begin{center}
			\includegraphics[width=0.9\textwidth]{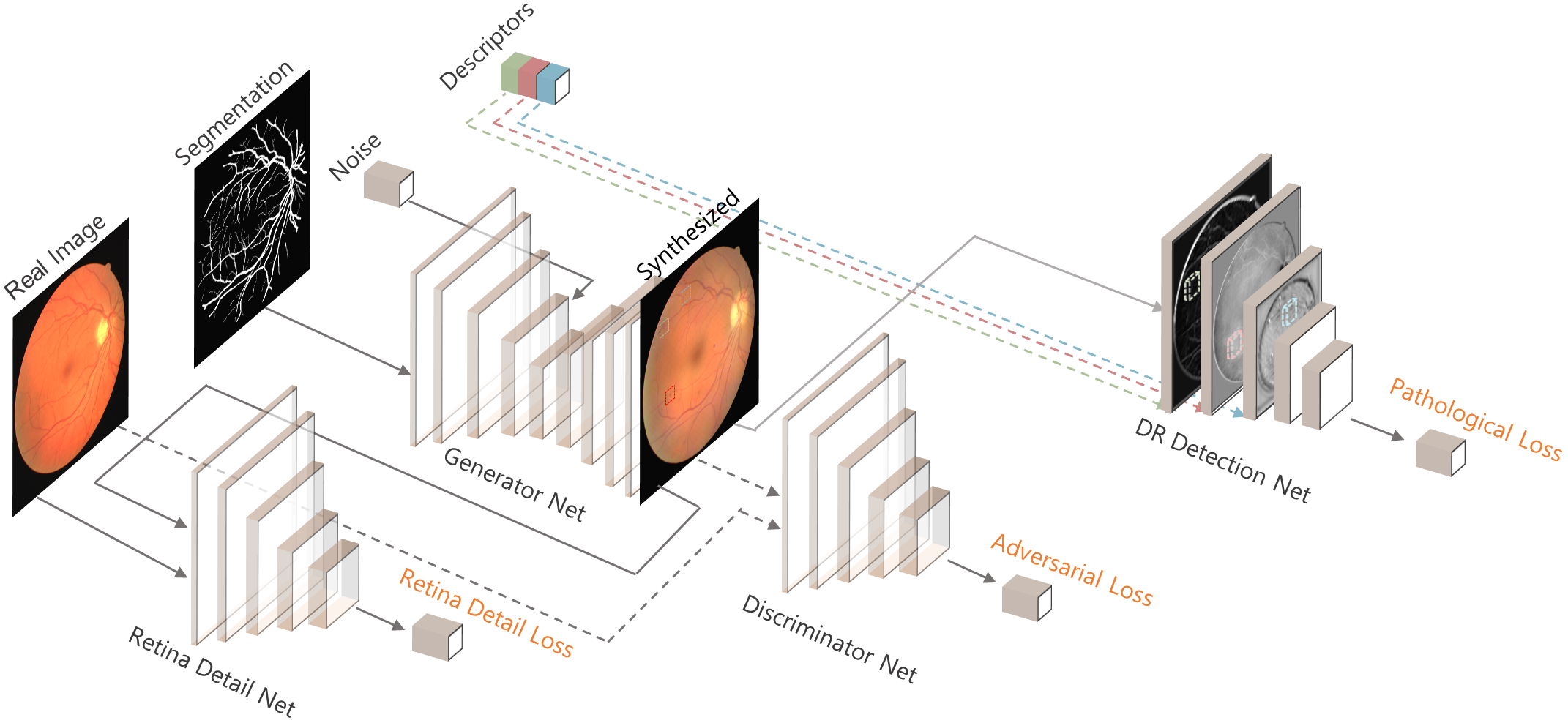}
		\end{center}
		\caption{The architecture and data flow of our symptom transfer GAN, which contains four nets in training phase. After training, the generator itself is able to synthesize retinal fundus with lesions on specific locations.}
		\label{fig:pipeline}
	\end{figure*}
	
	\section{Visualizing Pathological Descriptor}
    
    According to Koch's Postulates, though pathogen is invisible (at least for naked eye), its property could be observed on the subject after injecting the purified pathogen. Similarly, we evaluate and visualize the interpretative medical meaning of this descriptor by using a GAN based method to generate fully controllable DR fundus images.  Our goal is to synthesize the diabetic retinopathy fundus images(Fig~\ref{fig:koch}.(d)) that carry the lesions that appear on a pathological reference one (Fig~\ref{fig:koch}.(a)). Since our descriptor is lesion based, we could even arbitrarily manipulate the number and position of symptom. As shown in Fig~\ref{fig:pipeline}, with given descriptors $ \mathcal{D} $, we design a novel conditional GAN to achieve this.
    
    Our whole network structure consists of four sub-nets: the generator net, the discriminator net, the retina detail net, and the DR detection net. Given the vessel segmentation image and a noise code as input, the generator tries to synthesize a tubular structured phantom. The discriminator net tries to distinguish the synthesized images from the real ones. To further enhance the physiological details during generation, we use the retina detail net to constrain the detail reconstruction. The DR detection net is the \emph{key part} of our proposed architecture, which constrains the synthesized images with the user-specified pathological descriptors in feature level. After training, one can easily obtain synthesized fundus from vessel segmentations using the generator net.

	
	

	
	
	\subsection{Generator and Discriminator}
    
    We use a U-Net~\cite{ronneberger2015u} network structure for our generator network.
	Taking a segmentation image $y \in \{ 0,1 \}^{W \times H}$ with a noise code $z \in \mathbb{R}^Z $ as input, the network outputs a synthesized diabetic retinopathy fundus RGB image $\hat{x} \in \mathbb{R}^{W \times H \times 3}$. 	
	The entire image synthesis process can be expressed as $G_{\theta}$ : $ ( y,z ) \mapsto \hat{x}$. Similarly, we can also define discriminant function $D_{\gamma}$ : $ ( X , y) \mapsto p \in [0,1] $. When $X$ is the real image $x$, $p$ should tend to 1 and when $X$ is the composite image $\hat{x}$, $p$ should tend to 0.	
	We follow the GAN's strategy and solve the  following optimization problem that characterizes the interplay between $ G $ and $ D $:
	\begin{align}
	\label{align_allloss} \textstyle
	\max_{\theta} & \textstyle\min_{\gamma} L(G_{\theta}, D_{\gamma}) = \nonumber \\
	&\mathbb{E}_i [L_\mathrm{adv}(i, \theta, \gamma) + L_\mathrm{retina}(i, \theta) 
	  + L_\mathrm{patho} (i, \theta)],
	\end{align}
	where $
	L_\mathrm{adv} = \log D_\gamma(x_i, y_i) + 
	 \log(1 - D_\gamma(G_\theta(y_i,z_i),y_i)),
	$ is the adversarial loss, with $ L_\mathrm{retina} $ and $ L_\mathrm{patho} $ being retina detail loss and pathological loss.
	To be more specific, learning the discriminator $ D $ amounts to maximizing $ -L_D = L_\mathrm{adv} $
	and the generator $ G $ is learned by minimizing a loss $ L_G = \tilde L_\mathrm{adv} + L_\mathrm{retina} + L_\mathrm{patho} $ with a simpler adversarial
	\begin{align}
	\label{align_Gloss} \textstyle
	\tilde L_\mathrm{adv} = -\log D_\gamma (G_\theta(y_i, z_i),y_i) .
	\end{align}
	
	
	
	
	
	
	
	
	
	\subsection{Retina Detail Loss}
    \label{sec:retina_detail}
    
    Though a  L1 loss (or MAE) between synthetic image could deliver a satisfactory result for style transfer application on common images, it fails to preserve the physiological details in the fundus. We will elaborate this in the experiment section. Therefore, we define \emph{retina detail loss} as:
	\begin{align}
	\label{align_retina_loss} 
	L_\mathrm{retina} = w_\mathrm{dd} L_\mathrm{dd} + w_\mathrm{tv} L_\mathrm{tv},
	\end{align}
	where diverge of details $L_\mathrm{dd}$ is meant to preserve physiological details in the fundus, while total variance loss $L_\mathrm{tv}$ is the global smoothing term.
	
		We choose to measure diverge of details in VGG-19~\cite{DBLP:journals/corr/SimonyanZ14a} feature space. For specific layer $ \lambda $ and VGG feature extraction function $ F_V^\lambda $,
	\begin{align}
	\label{align_dd_loss} 
	L_\mathrm{dd} =  {\lVert F_V^\lambda(x_i) - F_V^\lambda(\hat x_i) \lVert}.
	\end{align}
	
	In addition, to ensure the overall smoothness, we also regulate image gradients to encourage spatial smoothness $L_\mathrm{tv}$:
	\begin{align}
	\label{align_tv_loss} \textstyle
	\sum_{w,h}{\lVert \hat{x}_i^{(w,h+1)} - \hat{x}_i^{(w,h)} \lVert}
	+ {\lVert \hat{x}_i^{(w+1,h)} - \hat{x}_i^{(w,h)} \lVert}.
	\end{align}
	
	\subsection{Pathological Loss}
   
   To constrain the synthesized image to carry pathological features, we enforce it to have the similar detector neuron activation on lesion regions to the reference image. In this way, the synthesized image should be equivalent to the reference from DR detector point of view. We regulate the neuron activation to be close to the ones recorded in pathological descriptors $ \mathcal{D} $ with a  \emph{pathological loss} :
    \begin{align}
	\label{align_patho_loss} 
	L_\mathrm{patho} = w_\mathrm{dp} L_\mathrm{dp} + w_\mathrm{mv} L_\mathrm{mv},
	\end{align}
	where the pathological diverge $L_\mathrm{dp}$ is the differences on features summed over lesion regions and layers, while masked variance loss $L_\mathrm{mv}$ represents the local smoothing term. 
    
    As shown in  Fig~\ref{fig:pipeline}, we input the synthesized image $ \hat x_i $ into the pre-trained DR detector.
    To ensure the extracted feature pathes $ \mathcal{F}_{l\rho} $ in fixed pre-specified regions $ \rho(d) $ are close to those $ F_{lr} $ in given descriptors $ d \in \mathcal{D} $ across all layers $ l \in \Lambda $, we define $ L_\mathrm{dp}$ as ($ \otimes $ means elementwise multiply)
    \begin{align}
    \label{align_dp_loss} \textstyle
    L_\mathrm{dp} = \frac{1}{|\mathcal{D}|} \sum_{d\in \mathcal{D}} &\textstyle \frac{1}{|\Lambda|} \sum_{l \in \Lambda}\nonumber \\ \textstyle \frac{w_\mathrm{gram}}{W_\rho H_\rho} \cdot 
    || \mathbb{G}(\mathcal{M}_{ld}& \otimes  \mathcal{F}_{l\rho}(\hat x_i)) 
    - \mathbb{G}(\mathcal{M}_{ld} \otimes F_{lr}(d)) ||,
    \end{align}
	where $ \mathcal{M}_{ld} = M_{lr} \otimes \mathrm{normalize}(|A_{lr}|) $ is computed with $ M_{lr}, A_{lr}$, the elements of $ d_l $, and used as a feature mask. The binary mask $ M_{lr} $ restricts loss in pixel-level, and $ A_{lr} $ stresses the lesion region as a soft mask.
	In this definition, we measure the diverge by the symmetric Gram matrix $ \mathbb{G}_{K\times K} $, which represents the covariance of different channels in feature maps $ F \in \mathbb{R}^{W\times H\times K}$:
	\begin{align}
	\label{align_gram} \textstyle
	\mathbb{G}(F)_{i, j} = \sum_{w,h} F_{whi}F_{whj}.
	\end{align}
	
	At the same time, our network would integrate the synthetic lesion features into the current background. First of all, a lesion redistribution mask $ M_\mathrm{rd} $, which covers lesion regions in the synthesized image, is computed based on all $ \rho $ and $ M_{0r} $. Then the mask is dilated to $ M_\mathrm{grd} $ by a gauss kernel to softly expand the boundary. With masked synthetic image $ \tilde{x}_i = M_\mathrm{grd} \hat{x}_i $, we define masked variance loss $L_\mathrm{mv}$ as
	\begin{align}
	\label{align_mtv_loss} \textstyle
	\sum_{w,h}{\lVert \tilde{x}_i^{(w,h+1)} - \tilde{x}_i^{(w,h)} \lVert}
	+ {\lVert \tilde{x}_i^{(w+1,h)} - \tilde{x}_i^{(w,h)} \lVert}.
	\end{align}

	%

	\subsection{Implementation Details}

	The chosen norm in above equations is L1.  Weights for different losses are $ w_\mathrm{dd} = 1, w_\mathrm{tv} = 100, w_\mathrm{dp} = 10, w_\mathrm{mv} = 5 w_\mathrm{tv}, w_\mathrm{gram} = 10^6 $. Based on experience, we set $ \lambda $ to be the second convolutional layer in the fourth block of VGG. 
	
	The batch size is set to 1. Before each training step, the input image values are scaled to $[-1, 1]$, and a random rotation is performed on the input. 
	The training is done using the ADAM optimizer~\cite{kingma2014adam} and the learning rate is set to 0.0002 for the generator and 0.0001 for the discriminator. In order to ensure that generator and discriminator are adapted, we update generator twice then update discriminator once. During training, the noise code is sampled element-wise from zero-mean Gaussian with standard deviation 0.001; At testing run, it is sampled in the same manner but with a different standard deviation of 0.1.
	The training finishes after 20000 mini-batches. In addition, we find the result more robust if the whole model is initialized to a trained one with no pathological loss.

	\section{Experiment Results}
    	
	\subsection{Dataset and preparation}
	In this paper, we select three datasets: DRIVE~\cite{staal:2004-855}, STARE~\cite{hoover2000locating} and Kaggle~\cite{kaggle2016diabetic}. 
	DRIVE contains 20 training images and 20 test images, with each of size 584$\times$565$\times$3. STARE contains 40  images and the size is 700$\times$605$\times$3. The Kaggle dataset contains 53576 training images and 35118 test images of various size. The DR detector is trained on Kaggle dataset following ~\cite{oO2016detector}. When training generator, we uses binary image and its corresponding retinal image in the DRIVE dataset. Before processing, we change all of the image into size of 512 $\times$ 512$\times$3 following ~\cite{zhao2018synthesizing}.
	
    
    \subsection{Visualization of Pathological Descriptor}
    
\begin{figure}[t]
		\begin{center}
			\includegraphics[width=\columnwidth]{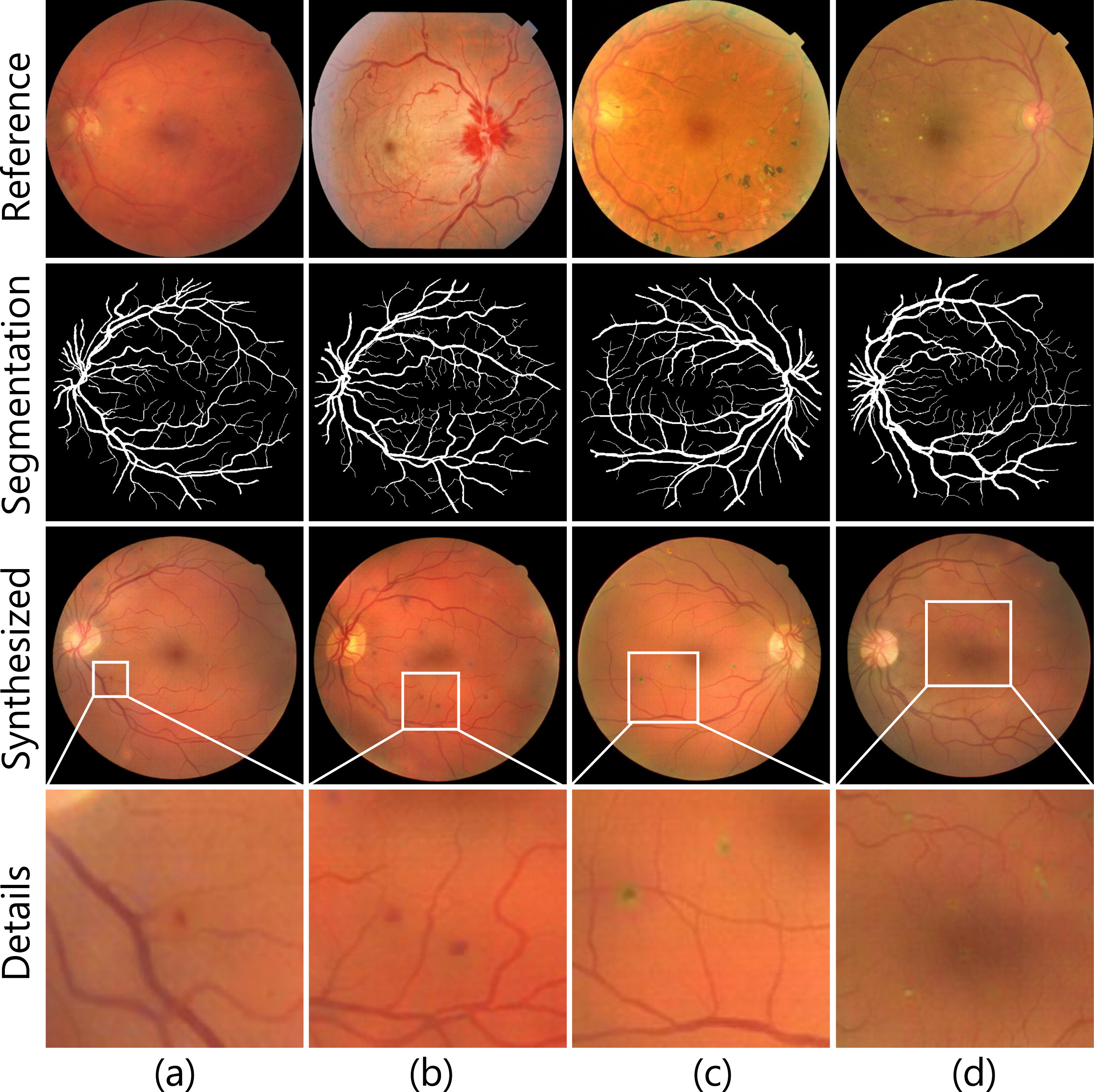}
		\end{center}
		\caption{Results of our experiment. We use reference images in row 1 and vessel segmentations in row 2, to generate synthesized retinal fundus, as shown in row 3-4.}
		\label{fig:results_show}
	\end{figure}

    We use images in STARE and Kaggle as pathological references to extract the pathological descriptor. The position of individual lesion is randomly chosen.  We have organized a group of 5 ophthalmologists to evaluate our results. 
    After training the generator, we test it on binary vessel images from DRIVE test set. The exemplary result in Fig~\ref{fig:results_show} shows that our pathological descriptors contain appearance and color features of lesions in different types. Microaneurysms in (a) and (b) looks very plausible.  The laser scars in (c) are consistent with the fundus of treated RD patients. However, our generated hard exudates in (d) are of some artifacts.
     
\begin{figure}[t]
		\begin{center}
			\includegraphics[width=\columnwidth]{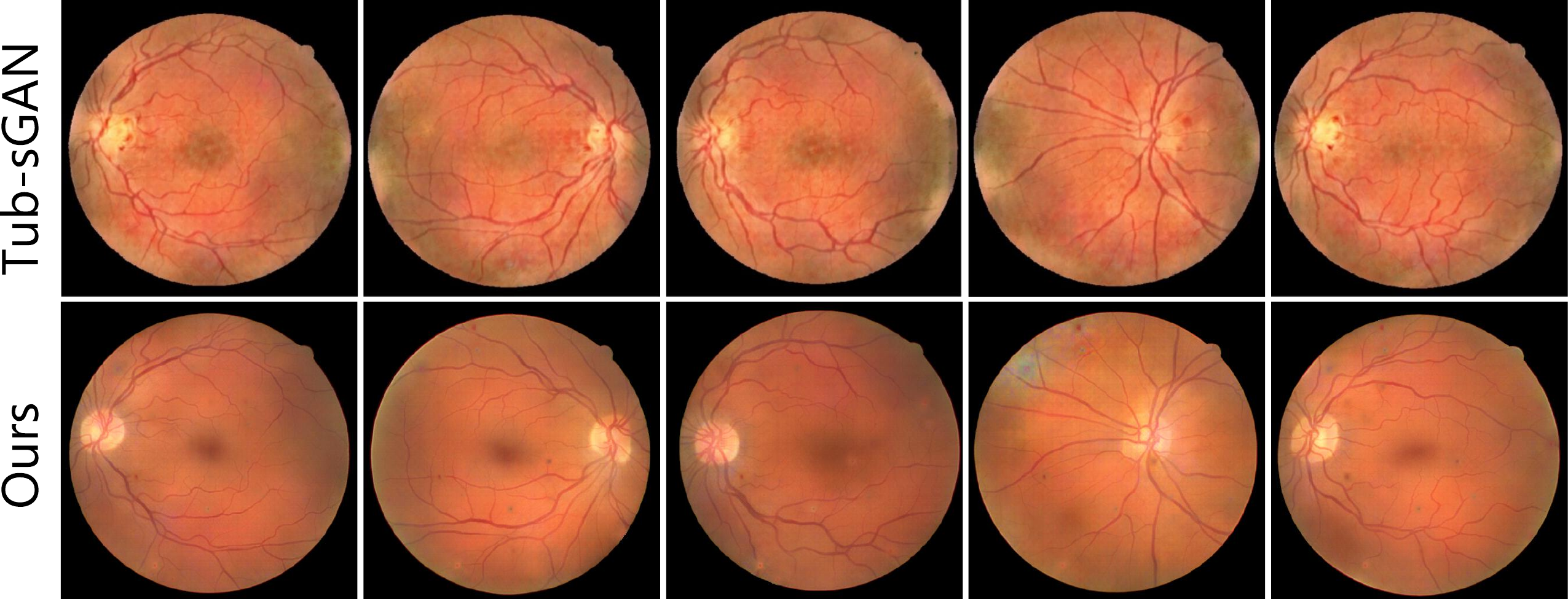}
		\end{center}
		\caption{Pathological details comparison between Tub-sGAN and ours. }
		\label{fig:compare_tubsgan}
	\end{figure}

    In Fig~\ref{fig:compare_tubsgan}, we compare Tub-sGAN's results with ours. Our method generates images with realistic lesion details where Tub-sGAN fails. For example, in images synthesized by our method (row 2), we can clearly spot the microaneurysms appeared at the far end of vessels. However, the lesions of Tub-sGAN (row 1) could not be classified into any known symptom. In addition, our method could output images with clear vessels, and the optic disc is better with concave appearance.
    
	\subsection{Quantitative Comparison}
	
	To further strengthen our method, we organized a peer review by a board of 5 professional ophthalmologists. The ophthalmologists were asked to double-blindly evaluate the randomly shuffled fundus images synthesized by our method and Fila-sGAN. For each image, they gave three scores (ranged 1-10, higher indicates better): 1. realness of the fundus image, 2. realness of the lesions, 3. severity of the DR. Finally, we collected valid scores on 560 images, of which the average scores are show in Table \ref{tab:quantitative_comparison}. The p value of T-test is 9.80e-10 and 7.95e-5 for fundus and lesion realness respectively that our mean score is higher than Fila-sGAN.
	
	\begin{table}[h!]
		\begin{center}
			\begin{tabularx}{0.8\columnwidth}{Xccc}
				\hline
				& Score 1 & Score 2 & Score 3 \\ \hline
				Tub-sGAN & 2.91 & 2.53 & 2.53 \\ 
				Ours & 4.21 & 3.37 & 3.08 \\ \hline
			\end{tabularx}
		\end{center}
		\caption{Average scores from the ophthalmologists.}
		\label{tab:quantitative_comparison}
	\end{table}
	
    \subsection{Lesion Manipulation}
        	
\begin{figure}[h!]
		\begin{center}
			\includegraphics[width=\columnwidth]{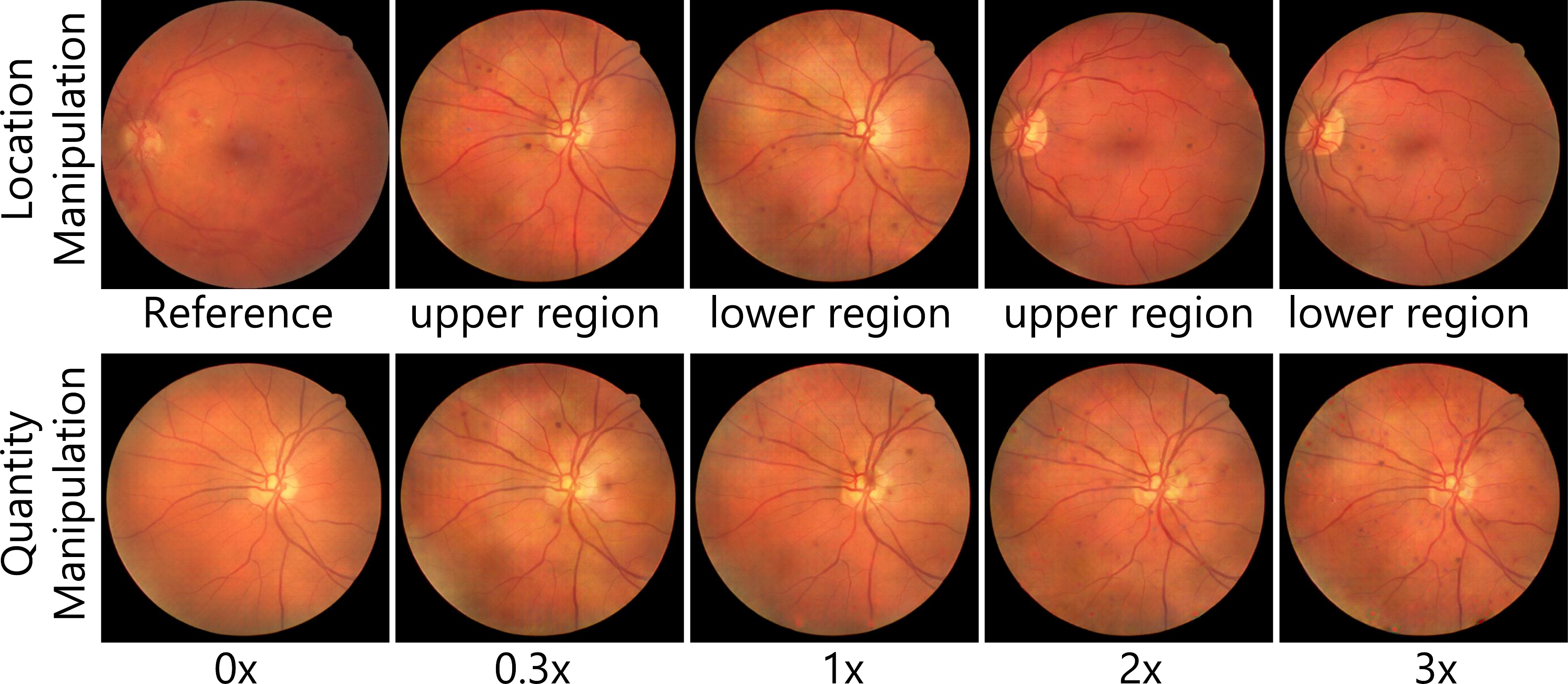}
		\end{center}
		\caption{Results of lesion manipulation.}
		\label{fig:redistribute}
        \end{figure}

	As mentioned above, our method is lesion-based, which makes lesion-wise manipulation possible. As shown in Fig~\ref{fig:redistribute}, we trained two generators from the same reference image (row 1, col 1) but distribute the lesion to the upper region (row1, col 2\&4) and lower region(row1, col 3\&5) respectively. On the other hand, we can also control the number of lesions. For example, drop out some of descriptors to generate less lesions, or clone some of descriptors to get more lesions. Row 2 in Fig~\ref{fig:redistribute} shows a series of synthesized pictures with lesion number increasing from zero to 3 times of that in the reference image, and the varying severity of the DR symptom could be observed in the fundus images.

\begin{figure}[t]
		\begin{center}
			\includegraphics[width=\columnwidth]{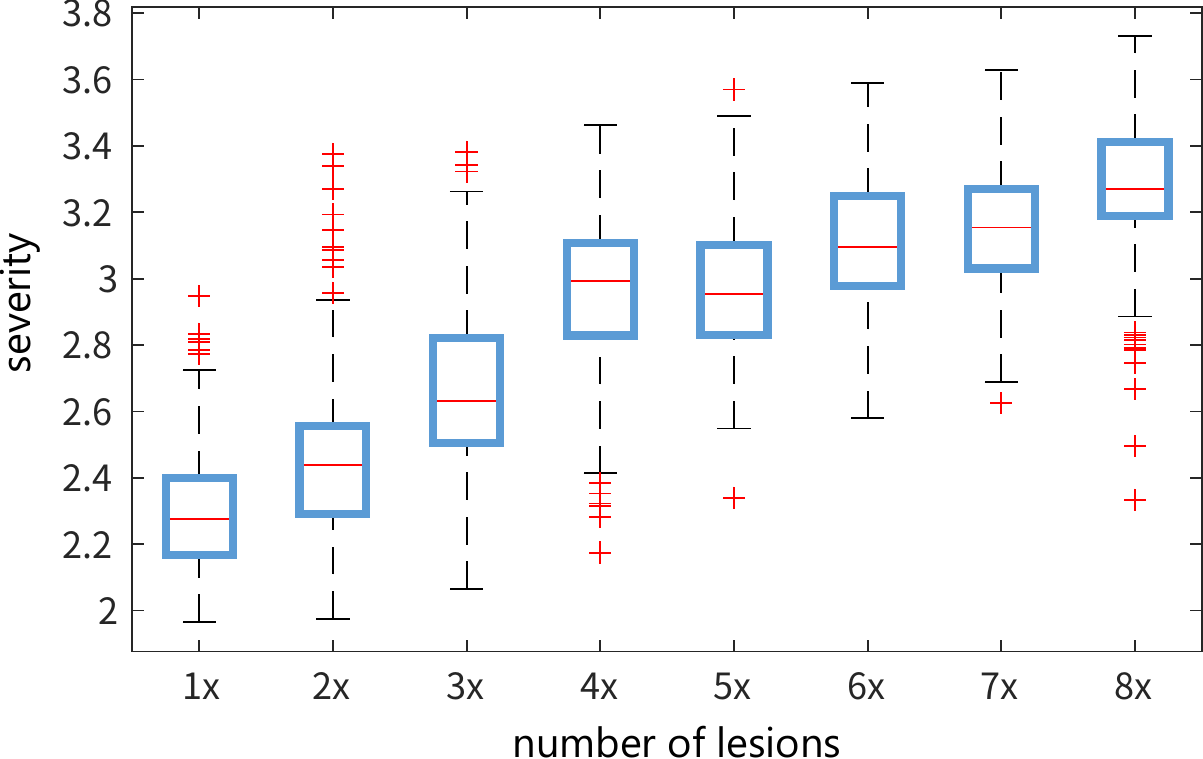}
		\end{center}
		\caption{The severity score with increasing lesions number.}
		\label{fig:number_severity}
	\end{figure}

  We also evaluate the above synthesized images on the Diabetic Retinopathy Detector~\cite{oO2016detector}.  According to~\cite{aao2002drscale}, number of microaneurysms is an important criteria for the severity diagnose.  Here we manipulate the number of microaneurysms and count its severity prediction. For each lesion number, we synthesize 10 images on each segmentation in DRIVE test set which has 20 testing image. Thus, we count the predicted severity for 200 images over each number of lesions. As reported in Fig~\ref{fig:number_severity}, by manipulating the lesion number, we receive consistency result from DR detector.
  
    \subsection{Detail Preservation}
    \label{sec:exp_detail}
    
\begin{figure}[h!]
		\begin{center}
			\includegraphics[width=\columnwidth]{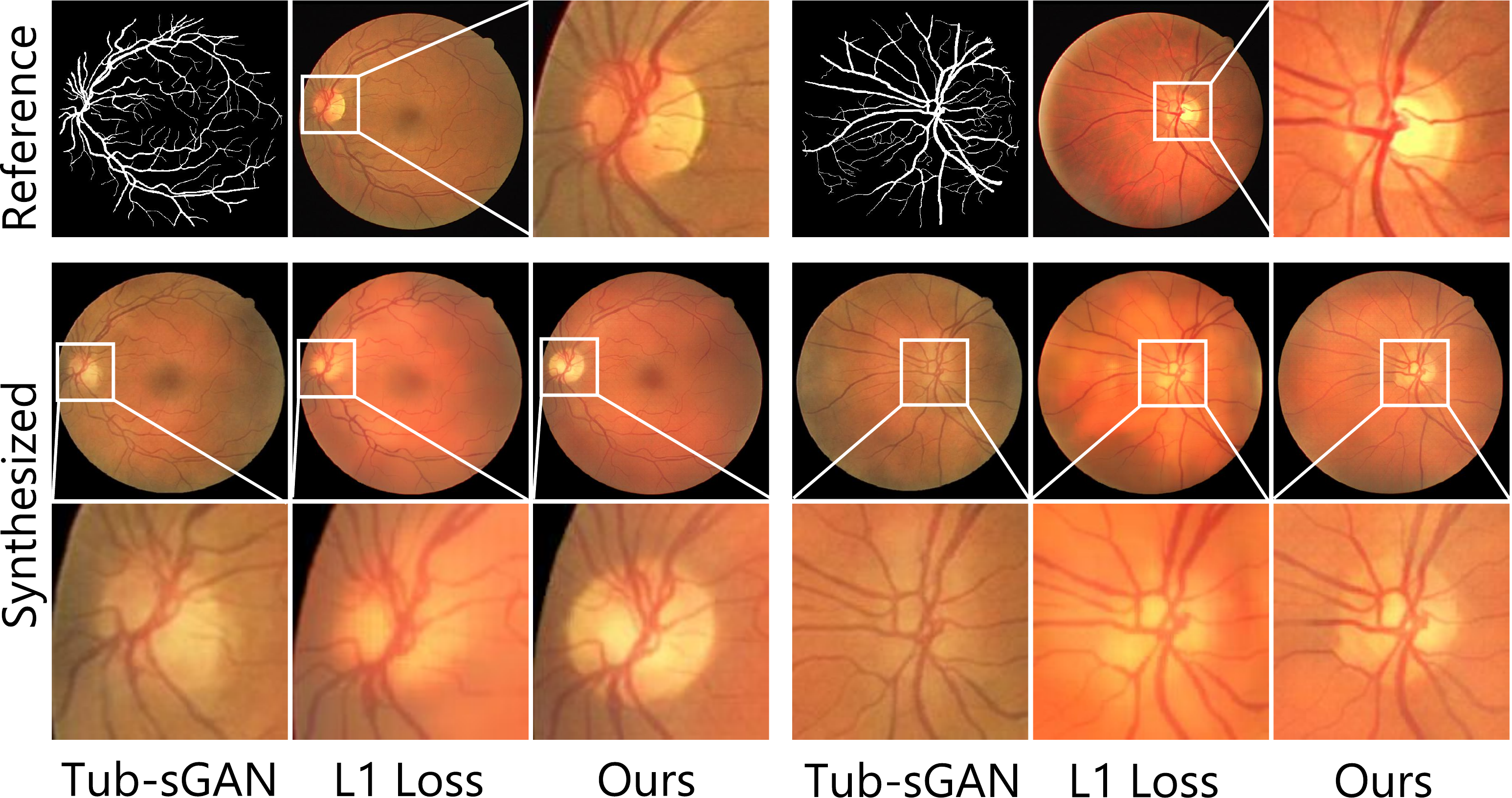}
		\end{center}
		\caption{The bright area of optic disc. Row 1 are two real fundus reference images with a clear boundary around each optic disc. Below are fundus synthesized by different methods, among which ours is best in realism. }
		\label{fig:optic_disc}
	\end{figure}

    As we discussed above, it is intuitive to regulate the difference between real and synthesized images on L1 loss as most of style transfer application~\cite{DBLP:conf/cvpr/GatysEB16}. However, this kind of metrics such as MAE, MSE, and SSIM only focus on low-level information in the image. In our practice on retinal image, we find the images generated with L1 loss rather than our retina detail loss fail to preserve some important physiological details such as optic disc boundary and choroid. Here we compare the images generated by Tub-sGAN~\cite{zhao2018synthesizing}, method with L1  loss and our current application in Fig~\ref{fig:optic_disc}.   We can see our method with a retina detail loss is appropriate for synthesizing photorealistic fundus images.
    
	\subsection{Ablation study}
	
    We evaluate the effects of individual components by adjusting their weights. 
    
	
    \subsubsection{Retina Detail Loss}
    
\begin{figure}[t]
		\begin{center}
			\includegraphics[width=\columnwidth]{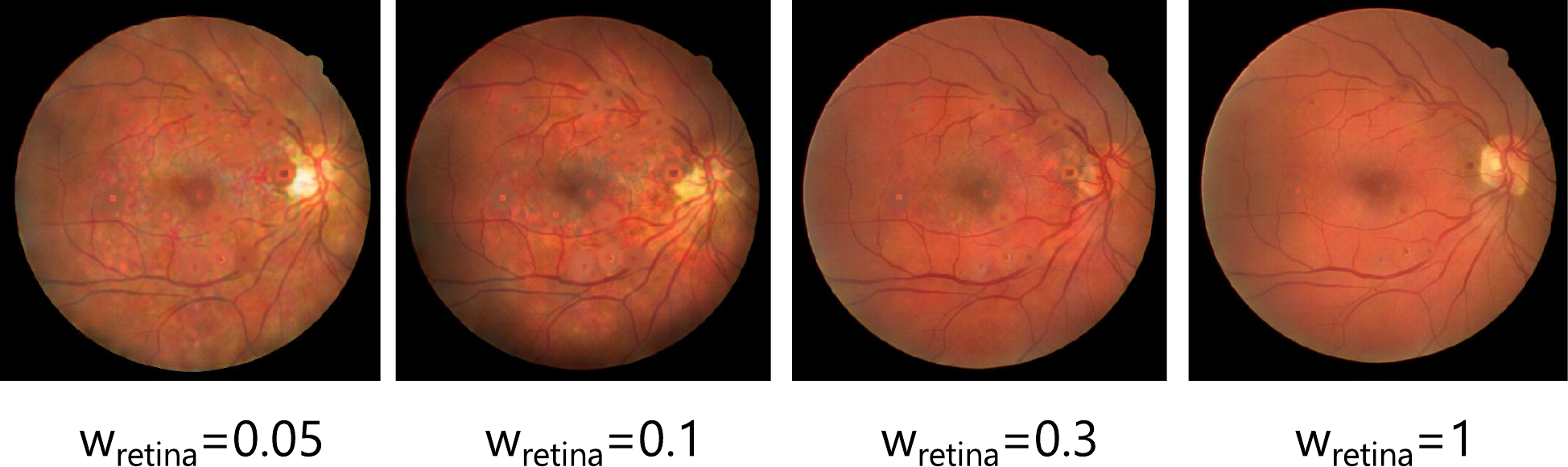}
		\end{center}
		\caption{The generated image with increasing weights of retina detail loss.}
		\label{fig:ablation_retina}
	\end{figure}

	The retina detail loss serves to preserve physiological detail. When reducing the weight of retina detail loss, the synthesized optic disc blurs, and noises in the image increase, as shown in Fig~\ref{fig:ablation_retina}.
    
    \subsubsection{Pathological Loss}
    
\begin{figure}[h!]
		\begin{center}
			\includegraphics[width=\columnwidth]{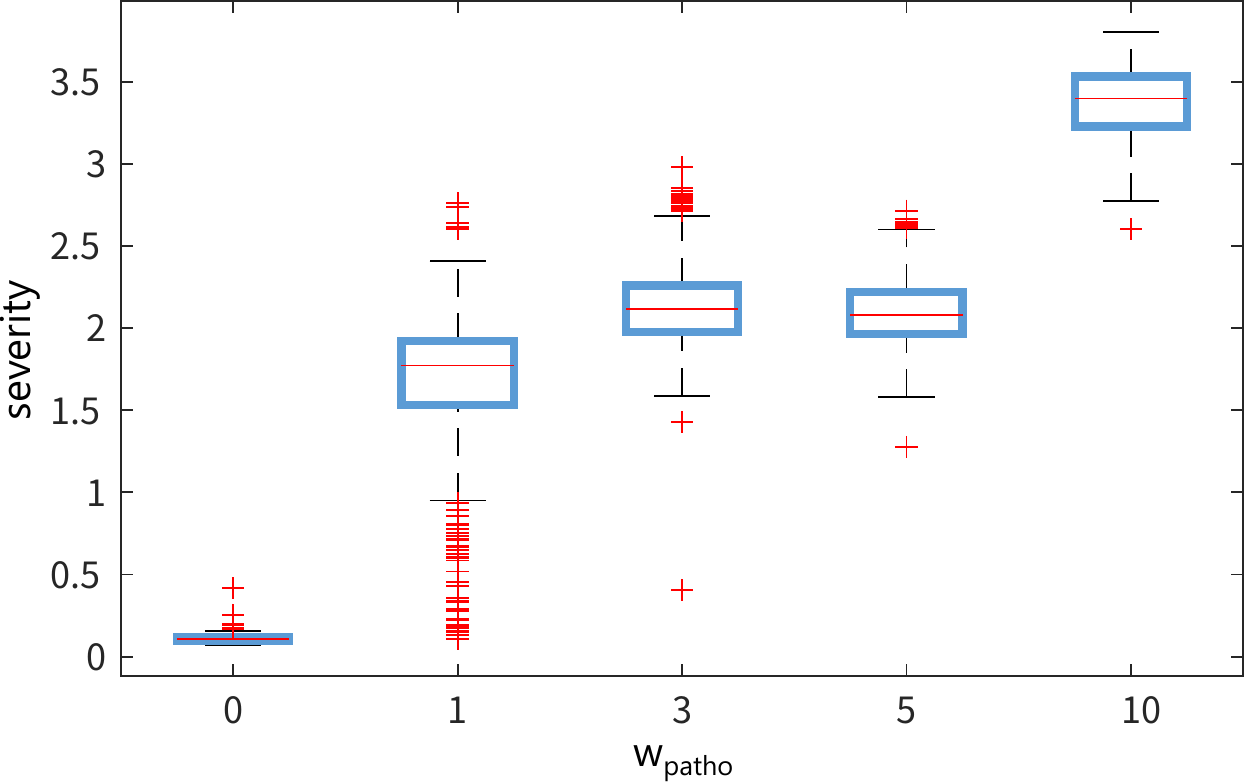}
		\end{center}
		\caption{The severity score with increasing weights of pathological loss.}
		\label{fig:ablation_patho}
	\end{figure}

    The pathological loss controls the synthesized lesions. When reducing the weights of pathological loss, we observe that lesions become weaker and weaker before they disappear. To further confirm this point, we evaluate the severity score of the generated images by DR detector. As Fig~\ref{fig:ablation_patho} shows, the severity increases with the weight of pathological loss. It worth pointing out that the severity score is $0$ when the constraints of pathological loss is absent. 
    
	
        \subsection{Color Transfer}
        
\begin{figure}[h!]
		\begin{center}
			\includegraphics[width=\columnwidth]{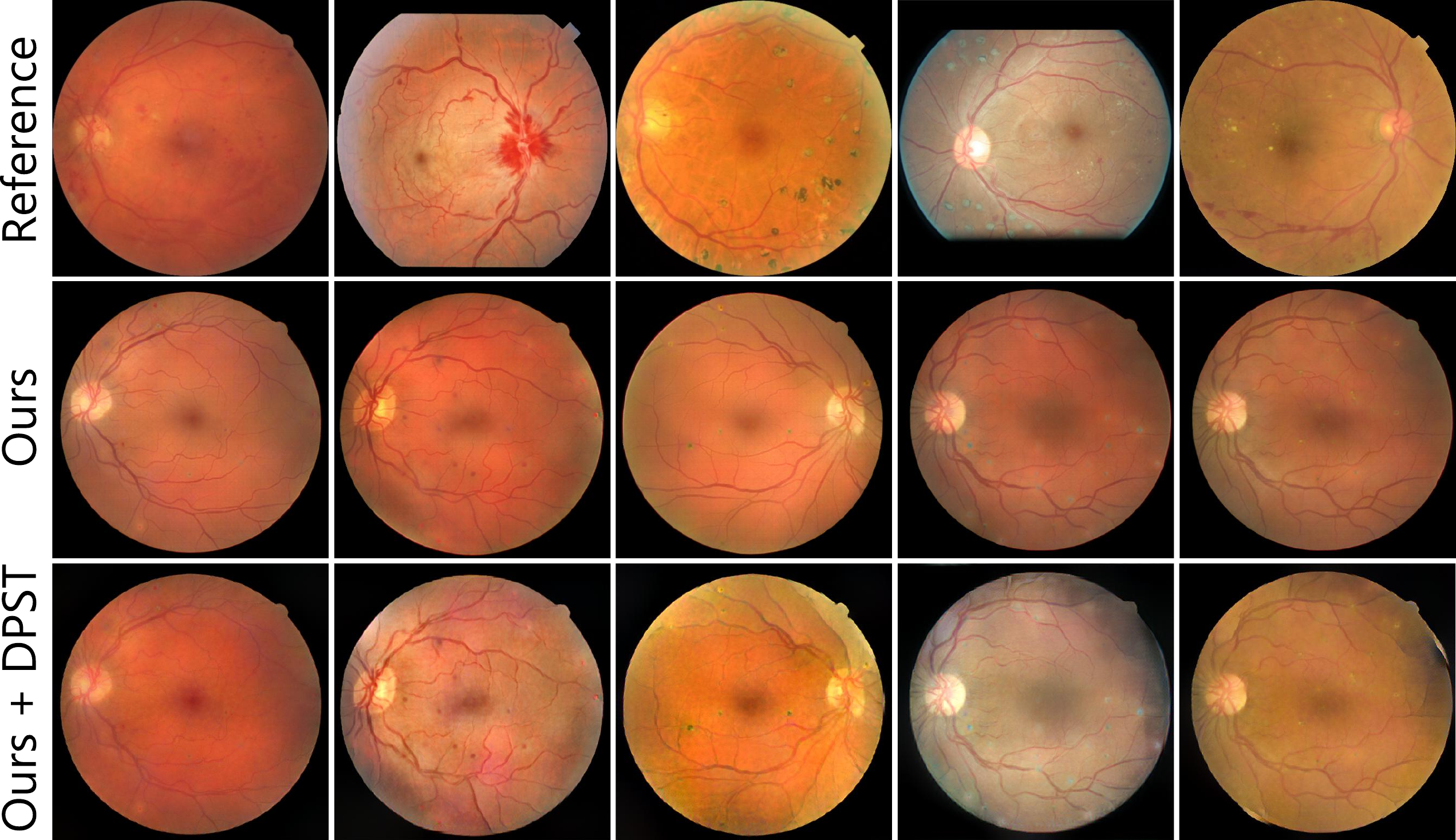}
		\end{center}
		\caption{Color transfer results. Our method with a post process achieves a better visual effect. }
		\label{fig:results_show_DPT}
	\end{figure}

        Unlike traditional style transfer,  our method focuses on transferring pathological feature rather than color or brightness. Here, we transfer the appearance by applying Deep Photo Style Transfer~\cite{DBLP:conf/cvpr/LuanPSB17} (DPST) after our synthesized image. As shown in Fig~\ref{fig:results_show_DPT}, compared to the direct synthesized image at row 2, the image after DPST possess a higher color consistency while preserving the  pathological lesions such as microneurysms.


    

    

	\subsection{Computation Time}

	All the experiments are tested out on a sever with Intel Xeon E5-2643 CPU, 256GB memory and Titan-Xp GPU. Training time on DRIVE and testing time are shown in table~\ref{tab:computation_time}. Compared to Tub-sGAN~\cite{zhao2018synthesizing}, we are faster which benefits from a more streamlined feature extraction network and descriptor-based comparison.
	
	\begin{table}[h!]
		\begin{center}
		\begin{tabularx}{\columnwidth}{XrrX}
			\hline
			 & Training & Testing & Platform \\ \hline
			Tub-sGAN & 108/109 min & 0.45/0.12 s & Titan-X/Xp \\ 
			Ours & 90 min & 0.12 s & Titan-Xp \\ \hline
		\end{tabularx}
		\end{center}
		\caption{Computation time of the different methods. The Tub-sGAN time on Titan-X is reported in their paper, and its time on Titan-Xp is mesured by us.}
		\label{tab:computation_time}
	\end{table}
		

	

	\section{Conclusion}
        To exploit the network interpretability in medical imaging, we proposed a novel strategy to encode the descriptor from the activated neurons that directly related to the prediction. To visually illustrate the extracted pathologic descriptor, we followed the similar methodology of Koch's Postulates that aim to identify the unknown pathogen. In addition, we proposed a GAN based visualization method to visualize the pathological descriptor into a fully controllable pathology retinal image from an unseen binary vessel segmentation. The retinal images we generated have shown medical plausible symptoms  as the reference image. Since pathological descriptor is associated with individual lesion and spatial independent, we could arbitrarily manipulate the position and quantity of the symptom. We verified the generated images with  a group of  licensed ophthalmologists and our result is shown to be both qualitatively and quantitatively superior to state-of-the-art. 
        The feedback of doctors shows our strategy  has strengthened their understanding on how deep learning makes prediction. Not limited in interpreting medical imaging, we will extend our strategy to more general interpretability problem.

        
        \subsubsection{Acknowledgement.} This work was supported by National Natural Science Foundation of China (NSFC) under Grant 61602020.


\bibliography{mybib}
\bibliographystyle{aaai}
\end{document}